\documentclass[10pt,twocolumn,letterpaper]{article}

\usepackage{cvpr}
\usepackage{times}
\usepackage{epsfig}
\usepackage{graphicx}
\usepackage{amsmath}
\usepackage{amssymb}
\usepackage{verbatim}
\usepackage{float}
\usepackage{caption}
\usepackage{subcaption}
\usepackage{stfloats}
\usepackage{pbox}
\usepackage{comment}

\usepackage[pagebackref=true,breaklinks=true,letterpaper=true,colorlinks,bookmarks=false]{hyperref}
\DeclareMathOperator*{\argmax}{argmax} 

\cvprfinalcopy

\ifcvprfinal\fi
\begin{document}

\newcommand{\MATTHIAS}[1]{{\bf\textcolor{red}{Matthias: #1}}}
\newcommand{\YAWAR}[1]{{\bf\textcolor{blue}{Yawar: #1}}}
\newcommand{\JULIEN}[1]{{\bf\textcolor{cyan}{Julien: #1}}}
\newcommand{\TODO}[1]{{\textcolor{red}{TODO: #1}}}

\newcommand{\OURS}{ViewAL}

\title{\OURS: Active Learning With Viewpoint Entropy for Semantic Segmentation}
%\author{Yawar Siddiqui\\
%Technical University of Munich\\
%{\tt\small yawar.siddiqui@tum.de}
%\and
%Julien Valentin\\
%Google\\
%{\tt\small julienvalentin@google.com}
%\and
%Matthias Nie{\ss}ner\\
%Technical University of Munich\\
%{\tt\small niessner@tum.de}
%}
\author{%
\hspace{-0.2cm}\parbox{4cm}{\centering Yawar Siddiqui$^{1}$}\quad
\parbox{4cm}{\centering Julien Valentin$^{2}$}\quad
\parbox{4cm}{\centering Matthias Nie{\ss}ner$^{1}$}\\[0.8em]
 $^{1}$Technical University of Munich\quad
 $^{2}$Google\quad
 \vspace{0.4cm}
}

%\author{\textbf{Yawar Siddiqui}\\
%Technical University of Munich\\
%\and
%\textbf{Julien Valentin}\\
%Google\\
%\and
%\textbf{Matthias Nie{\ss}ner}\\
%Technical University of Munich\\
%}

\twocolumn[{%
	\renewcommand\twocolumn[1][]{#1}%
	\maketitle
	\begin{center}
	    \vspace{-0.4cm}
		\includegraphics[width=0.95\linewidth]{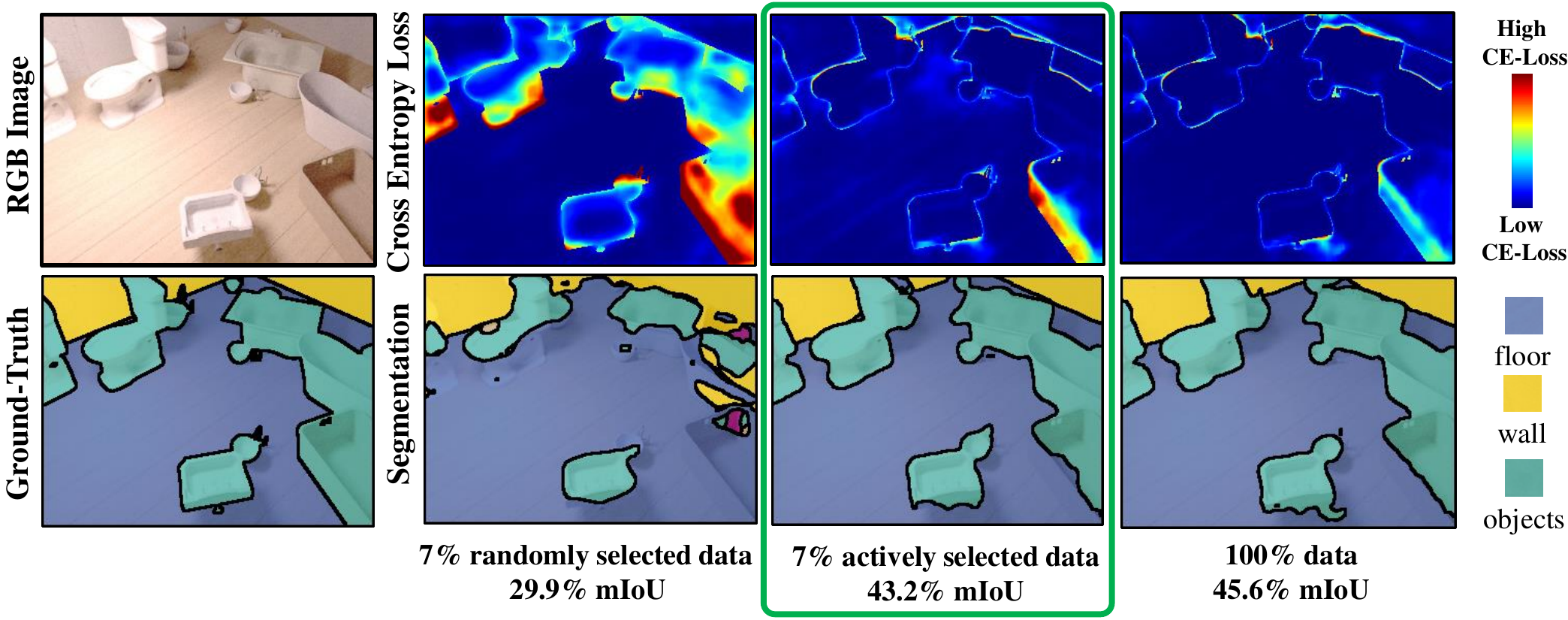}
		\captionof{figure}{ViewAL is an active learning method that significantly reduces labeling effort: with maximum performance attained by using 100\% of the data (last column), ViewAL achieves 95\% of this performance with only 7\% of data of SceneNet-RGBD~\cite{McCormac:etal:arXiv2016}. With the same data, the best state-of-the-art method achieves 88\% and random sampling (2nd column) yields 66\% of maximum attainable performance.}
		
		\label{fig:teaser}
	\end{center}    
	\vspace{0.5cm}
}]

\maketitle

% !TEX root = main.tex

\begin{abstract}
We propose \OURS{}
\footnote{Source code available: \href{https://github.com/nihalsid/ViewAL}{https://github.com/nihalsid/ViewAL}}
, a novel active learning strategy for semantic segmentation that exploits viewpoint consistency in multi-view datasets.
Our core idea is that inconsistencies in model predictions across viewpoints provide a very reliable measure of uncertainty and encourage the model to perform well irrespective of the viewpoint under which objects are observed.
To incorporate this uncertainty measure, we introduce a new viewpoint entropy formulation, which is the basis of our active learning strategy.
In addition, we propose uncertainty computations on a superpixel level, which exploits inherently localized signal in the segmentation task, directly lowering the annotation costs.
This combination of viewpoint entropy and the use of superpixels allows to efficiently select samples that are highly informative for improving the network.
We demonstrate that our proposed active learning strategy not only yields the best-performing models for the same amount of required labeled data, but also significantly reduces labeling effort.
Our method achieves 95\% of maximum achievable network performance using only 7\%, 17\%, and 24\% labeled data on SceneNet-RGBD, ScanNet, and Matterport3D, respectively. 
On these datasets, the best state-of-the-art method achieves the same performance with 14\%, 27\% and 33\% labeled data.
Finally, we demonstrate that labeling using superpixels yields the same quality of ground-truth compared to labeling whole images, but requires 25\% less time.

\end{abstract}

% !TEX root = main.tex

%%%%%%%%% BODY TEXT
\section{Introduction}

%par 1: DL needs AL for scaling
With the major success of deep learning on major computer vision tasks, such as image classification \cite{krizhevsky2012imagenet,sermanet2013overfeat,simonyan2014very,szegedy2015going}, object detection \cite{girshick2014rich,erhan2014scalable,girshick2015fast,ren2015faster}, pose estimation \cite{wei2016convolutional,pishchulin2016deepcut,insafutdinov2016deepercut,newell2016stacked}, or semantic segmentation \cite{long2015fully,ronneberger2015u,badrinarayanan2017segnet,chen2017deeplab,zhao2017pyramid}, both network sizes and the amount of data required to train these networks has grown significantly. 
This has led to a drastic increase in the costs associated with acquiring sufficient amounts of high-quality ground truth data, posing severe constraints on the applicability of deep learning techniques in real-world applications.
Active learning is a promising research avenue to reduce the costs associated with labeling. 
The core idea is that the system being trained actively selects samples according to a policy and queries their labels; this can lead to machine learning models that are trained with only a fraction of the data while yielding similar performance.
Uncertainty sampling is one of the most popular strategies in active learning to determine which samples to request labels for~\cite{wang2016cost,gal2017deep,beluch2018power,mackowiak2018cereals,joshi2009multi,tong2001support,seung1992query}.
Here, the model prefers samples it is most unsure about, based on an uncertainty measure, in order to maximize model improvement.
Existing uncertainty sampling techniques almost exclusively operate on single images, which is surprising since many consumer-facing applications, such as robots, phones, or headsets, use video streams or multi-view data coming from 3D environments.
As a consequence, geometric constraints inherently present in the real world are largely ignored, but we believe these are particularly interesting for examining the quality of network predictions; i.e., the same surface point in a scene should receive the same label when observed from different view points.
\begin{figure}[h]
\begin{center}
   \includegraphics[width=\linewidth]{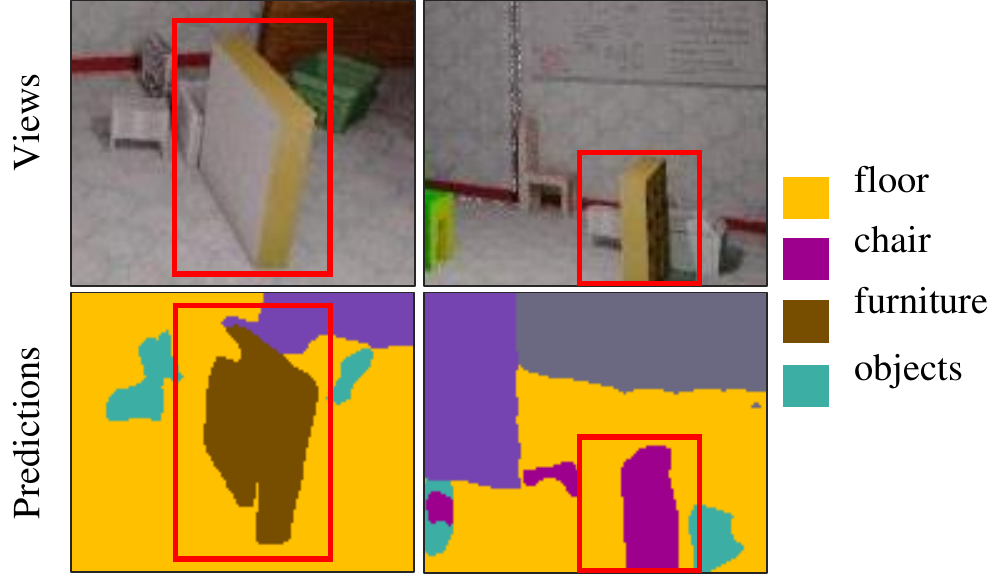}
\end{center}
\vspace{-0.4cm}
   \caption{Inconsistencies in segmentations for two views of the same object (\textit{furniture}). While in the first view the object is predicted to be \textit{furniture}, in the second view it is predicted to be a \textit{chair}.}
\label{fig:view_pred}
\end{figure}

In this work, we propose to exploit these constraints in a novel view-consistency-based uncertainty measure.
More specifically, we propose a new viewpoint entropy formulation which is based on the variance of predicted score functions across multiple observations.
If predictions for a given (unlabeled) object differ across views, we assume faulty network predictions; we then strive to obtain labels for the most uncertain data samples.
In addition, we propose uncertainty computations on a superpixel level, which exploits inherently localized signal in segmentation tasks, directly lowering the annotation costs. 
This combination of viewpoint entropy and the use of superpixels allows efficient selection of samples that are highly informative  for improving  the network.
Finally, we leverage sampling-based measures such as Monte Carlo (MC) dropout~\cite{gal2017deep}, and we show that in conjunction with these measures, we can further improve performance.
In summary, our contributions are: 
\begin{itemize}
    \item A novel active learning strategy for semantic segmentation that estimates model uncertainty-based on inconsistency of predictions across views, which we refer to as \textit{viewpoint entropy}.
    \item A \textit{most informative view} criteria based on KL divergence of prediction probability distributions across view points. 
    \item A superpixel-based scoring and label acquisition method that reduces the labeling effort while preserving annotation quality.
\end{itemize}

\section{Related Work}

%In this section, we will first touch upon the classic literature and then later introduce some recent DNN-based active learning approaches. 
 
A thorough review of classical literature on active learning can be found in Settles et al.~\cite{settles2009active}. As described in~\cite{yoo2019learning}, given a pool of unlabeled data, there are three major ways to select the next batch to be labeled: uncertainty-based approaches, diversity-based approaches, and expected model change. 
In uncertainty-based approaches, the learning algorithm queries for samples it is most unsure about. For a probabilistic model in a binary classification problem, this would mean simply choosing the samples whose posterior probability of being positive is nearest to 0.5~\cite{lewis1994sequential,lewis1994heterogeneous}. For more than two classes, entropy can be used as an uncertainty measure~\cite{settles2008analysis,hwa2004sample}. A simpler way is to select instances with the least confident posterior probabilities~\cite{settles2008analysis}. Another strategy could be to choose samples for which the most probable label and second most probable label have least difference in prediction confidence~\cite{joshi2009multi, roth2006margin}. Yet another uncertainty-based approach is querying by committee where a committee of multiple models is trained on the labeled data, and unlabeled samples with least consensus are selected for labeling~\cite{seung1992query,mccallumzy1998employing}.
 
Uncertainty-based approaches can be prone to querying outliers. In contrast, diversity-based approaches are designed around the idea that informative instances should be representative of the input distribution. Nguyen et. al~\cite{nguyen2004active} and Xu et al.~\cite{xu2007incorporating} use clustering for querying batches. 
The last method of expected model change~\cite{settles2008multiple,freytag2014selecting, kading2016active,Vezhnevets2012} queries samples that would change the current model most if their labels were known. It has been successful for small models but has seen little success with deep neural networks because of computational complexity involved. 

Quite a lot of the uncertainty-based approaches can be directly used with deep neural networks. Softmax probabilities have been used for obtaining confidence, margin, and entropy measures for uncertainty~\cite{wang2016cost}. Gal et al.~\cite{gal2017deep} use multiple forward passes with dropout at inference time (Monte Carlo dropout) to obtain better uncertainty estimates. Ensemble methods~\cite{beluch2018power, chitta2018large} have also been used to improve upon uncertainty estimates, however, these can be heavy in terms of memory and compute requirements. The loss learning approach introduced in~\cite{yoo2019learning} can also be categorized as an uncertainty approach. 
Sener et al.~\cite{sener2017active} propose a diversity-based approach, which formulates active learning as core-set selection - choosing a set of points such that a model learned over the selected subset is competitive for the remaining data points. Yang et al.~\cite{yang2017suggestive} present a hybrid approach, using both uncertainty and diversity signals. They utilize uncertainty and similarity information from a DNN and formulate sample selection as generalized version of the maximum set cover problem to determine the most representative and uncertain areas for annotation.

While most of these methods~\cite{wang2016cost,beluch2018power,sener2017active} have been verified on classification tasks, they can be easily adapted to target segmentation. Sun et al.~\cite{sun2015active} investigate active learning for probabilistic models (e.g Conditional Random Fields) that encode probability distributions over an exponentially-large structured output space (for instance semantic segmentation). Active learning for semantic segmentation with deep neural networks has been specifically investigated in~\cite{mackowiak2018cereals,yang2017suggestive,gorriz2017cost}. In~\cite{mackowiak2018cereals}, the authors use a regional selection approach with cost estimates that combine network uncertainty via MC dropout~\cite{gal2016dropout} with an effort estimate regressed from ground-truth annotation click patterns. 
View consistency for active learning for segmentation has not been investigated to the best of our knowledge. The work of~\cite{muslea2006active} comes close to ours in spirit, in which the authors investigate the effect of using multiple disjoint features (views), each of which describe the target concept. They show the effectiveness of using multiple views in active learning for domains like web page classification, advertisement removal, and discourse tree parsing. The work is extended in~\cite{zhang2009multi} for image classification tasks. % with simple models. 

% !TEX root = main.tex

\section{Method}

\begin{figure}[h]
\begin{center}
   \includegraphics[width=\linewidth]{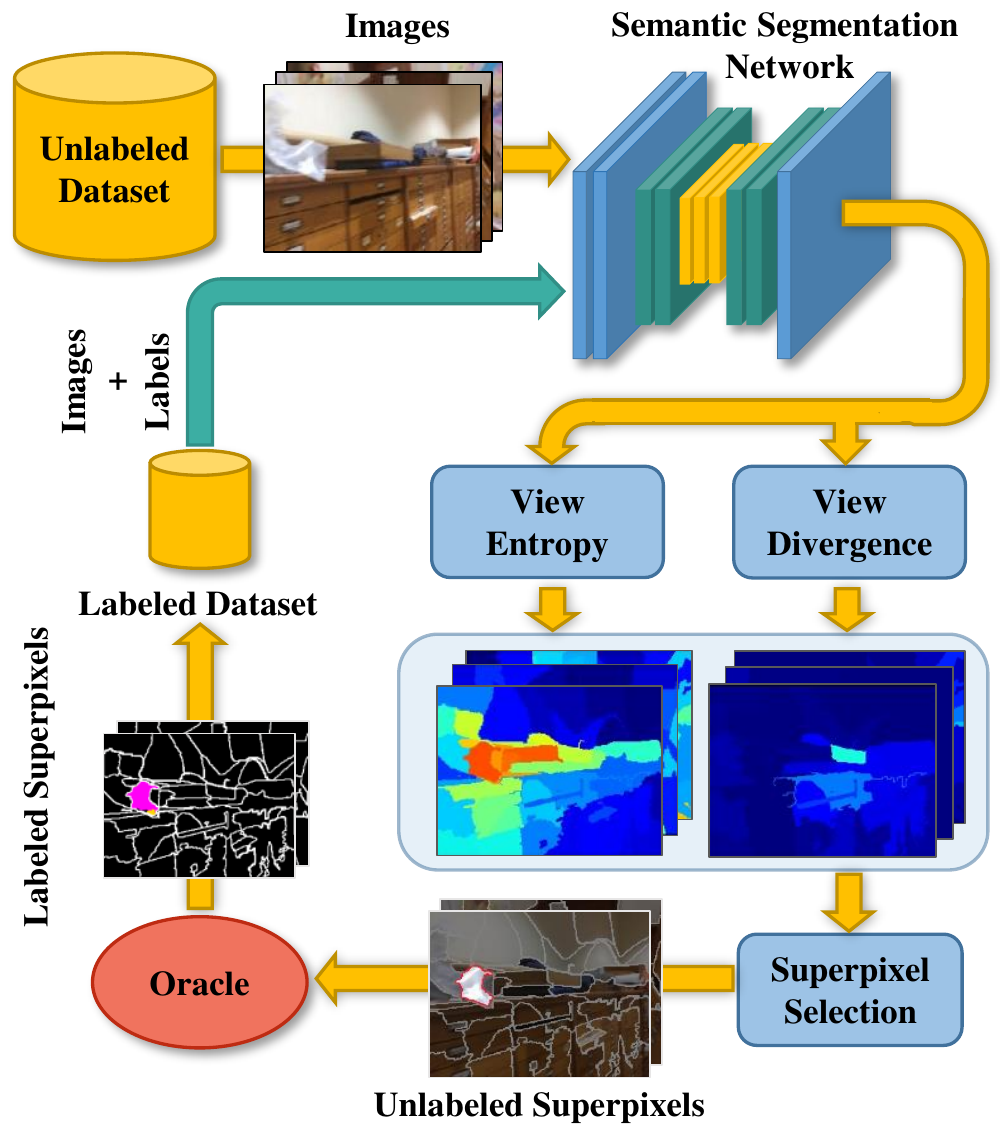}
\end{center}
\vspace{-0.44cm}
   \caption{Method overview: in each round of active selection, we first train a semantic segmentation network on the existing labeled data. Second, we use the trained network to compute a view entropy and a view divergence score for each unlabeled superpixel. We then select a batch of superpixels based on these scores, and finally request their respective labels from the oracle. This is repeated until the labeling budget is exhausted or all training data is labeled.}
\label{fig:overview}
\vspace{-7mm}
\end{figure}
\begin{figure*}
  \includegraphics[width=\textwidth]{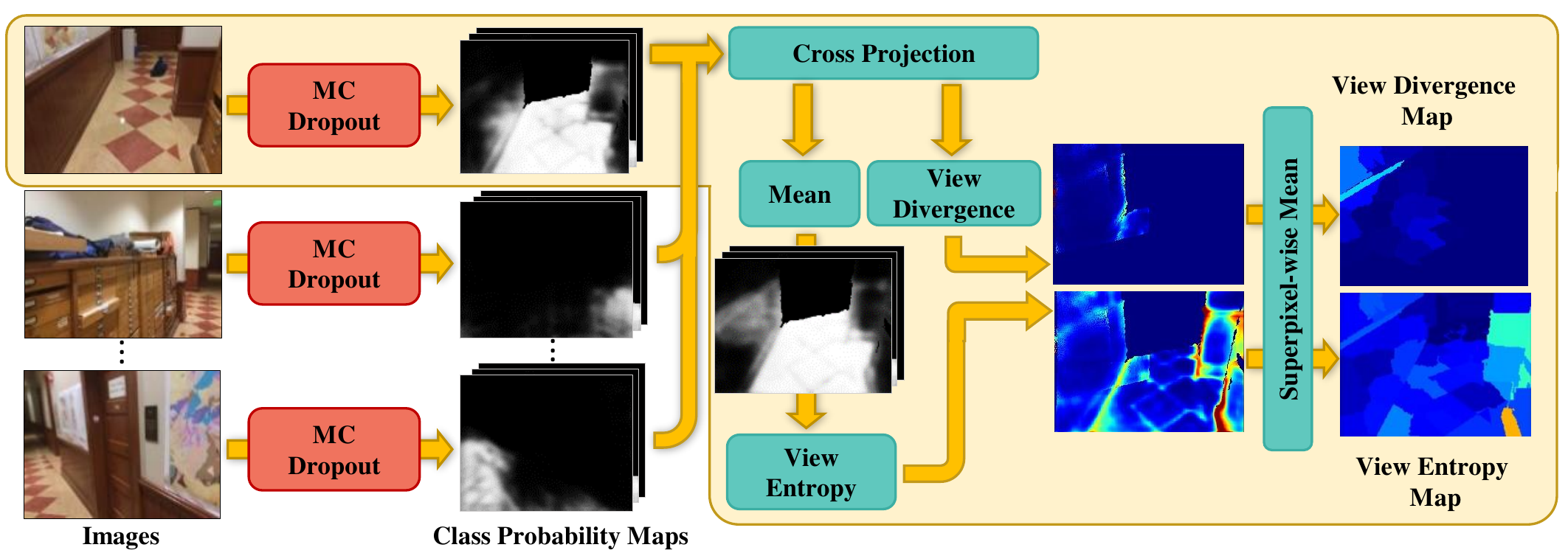}
  \caption{Computation of the view entropy and the view divergence scores. For each unlabeled superpixel in the dataset, we perform MC dropout~\cite{gal2016dropout} using 20 runs of dropout and average them to obtain class-probability maps. Next, we back-project each pixel and their associated class-probability distribution to 3D, and re-project all of these to the unlabeled superpixel, effectively providing multiple class-probability predictions per pixel. We then define the view entropy score as the entropy of the average class-probability distribution at each pixel. The view divergence corresponds to the average pairwise KL divergence between the class-distribution at any given pixel and the class-distributions projected at that pixel, effectively capturing the amount of agreement between the prediction in the current view with the prediction coming from other viewpoints. Finally, a view divergence score and a view entropy score is associated with each unlabeled superpixel by averaging the view divergence score and view entropy score of all the pixels they contain.}
  \label{fig:scorecalculation}
  \vspace{-3mm}
\end{figure*}

Our \OURS{} method consists of four main steps (see Fig.~\ref{fig:overview}): training the network on the current set of labeled data, estimating the model uncertainty on the unlabeled part of the data, selecting which super pixels to request labels for, and finally obtaining annotations. This series of steps is repeated until the labeling budget is reached or all the data labeled. We now describe these steps in more detail.

\subsection{Network Training}
We start by training a semantic segmentation network to convergence using currently labeled dataset $D_L$. Initially, $D_L$ is a small randomly selected subset of the dataset for which ground truth has been obtained. 

In theory, any semantic segmentation network can be used. We choose DeepLabv3+~\cite{chen2018encoder} with MobileNetv2~\cite{sandler2018mobilenetv2} as the backbone. We make this choice as DeepLabv3+ is one of the top performing segmentation networks on popular semantic segmentation benchmarks~\cite{pascal-voc-2012,cordts2016cityscapes}, and when combined with the MobileNetv2 backbone, it allows fast training, inference at low memory consumption. 

The MobileNetv2 backbone is initialized with weights from a model that was pre-trained on the ILSVRC \mbox{1000-class} classification task~\cite{russakovsky2015imagenet}. The rest of the layers use Kaiming initialization~\cite{he2015delving}. To prevent overfitting, we use blur, random crop, random flip, and Gaussian noise as data augmentations.

\subsection{Uncertainty Scoring}

Once the network is trained on $D_L$, our active learning method aims at predicting which samples from the unlabeled part of this dataset, $D_U$, are the most likely to be the most informative to the current state of the network. To this end, we introduce a new sample selection policy based on view entropy and view divergence scores. Fig.~\ref{fig:scorecalculation} provides an overview of these two new scoring mechanisms.

\subsubsection{View Entropy Score}
\label{section:view_entropy}
In a nutshell, the proposed view entropy score aims at estimating which objects are consistently predicted the same way, irrespective of the observation viewpoint. For each image, we first calculate its pixel-wise class probability maps using the segmentation network. To make the probability estimates more robust to changes in the input, we use the MC dropout method~\cite{gal2016dropout}. The probability for a pixel at position $(u, v)$ in image $I_i$ to belong to class $c$ is given by
\begin{equation}
\label{eqn:mcdropout_p}
    \mathit{P_{i}^{(u,v)}(c)} = \frac{1}{D}\sum_{d=1}^{D}{P^{(u,v)}_{i,d}(c)},
\end{equation}
where $D$ is the number of test time dropout runs of the segmentation network, and $P^{(u,v)}_{i,d}(c)$ is the softmax probability of pixel $(u,v)$ belonging to class $c$ in the MC dropout run $d$.

Next, using pose and depth information, all the pixels from the dataset and their associated probability distribution are back-projected to 3D, and projected onto all images. Each pixel $(u,v)$ in image $I_i$ is now associated with a set of probability distributions $\Omega^{(u,v)}_i$, each coming from a different view;

\begin{equation}
\label{eqn:distribution_set}
    \mathit{\Omega^{(u,v)}_i} = \{\mathit{P_{j}^{(x,y)}}, j \mid I_j(x,y)\text{ cross-projects to }I_i(u,v)\}
\end{equation}

The mean cross-projected distribution $Q_{i}^{(u,v)}$ can then be calculated as 
\begin{equation}
\label{eqn:view_marg}
    \mathit{Q^{(u,v)}_i} = \frac{1}{|\mathit{\Omega^{(u,v)}_i}|}\sum_{P \in {\mathit{\Omega^{(u,v)}_i}}} P^{(u,v)}
\end{equation}
which can be seen as marginalizing the prediction probabilities over the observation viewpoints. Finally, the view entropy score $\mathit{VE^{(u,v)}_i}$ for image $I_i$ is given by
\begin{equation}
    \mathit{VE^{(u,v)}_i} = -\sum_{c}\mathit{Q^{(u,v)}_i(c)} \log \mathit(Q^{(u,v)}_i(c))
\end{equation}

\subsubsection{View Divergence Score}
\label{section:view_divergence}
\begin{figure}[t]
	\begin{center}
    \includegraphics[width=\linewidth]{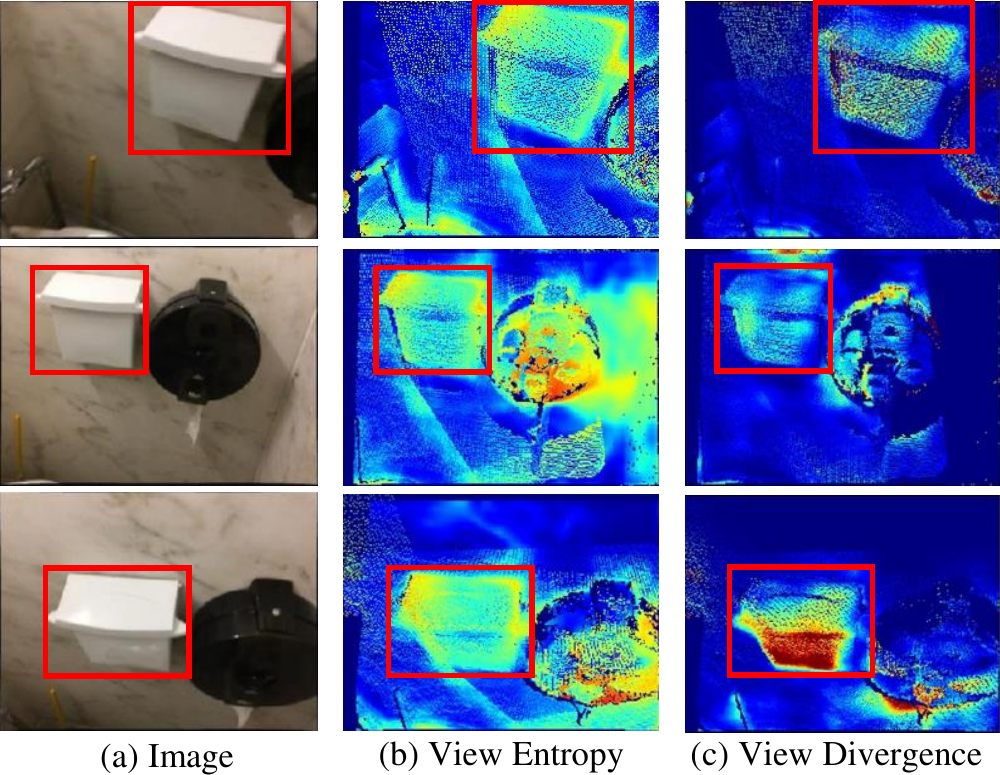}
    \end{center}
    \vspace{-0.4cm}
	\caption{View entropy and view divergence scores. For all score-maps, blue indicates low values, and red indicates high values. Pixels that are in correspondence hold the same view entropy score (b), since this score corresponds to a measure computed on the average class-probability of coming from all the pixels in correspondence. We then use the view divergence (c) to define which is the most disagreeing view, and send it to the oracle for annotation.}
	\label{fig:score_vis}
	\vspace{-2mm}
\end{figure}

Since the view entropy indicates for each pixel how inconsistent the predictions are across views, this score is the same for all the pixels that are in correspondence (Fig.~\ref{fig:score_vis}(b)) since it is calculated using probabilities marginalized over different views (Eq.~\ref{eqn:view_marg}). At this stage we are then able to establish the objects for which the network makes view inconsistent predictions, but we still need to determine which view(s) contains the largest amount of beneficial information to improve the network. To this end, we calculate a view divergence score for each pixel, which indicates how predictions about a particular 3D point observed in other views differ from the corresponding prediction in the current image. The view divergence score $\mathit{VD^{(u,v)}_i}$ for a pixel $(u,v)$ in image $I_i$ is given by
\begin{equation}
    \mathit{VD^{(u,v)}_i} = \frac{1}{|\mathit{\Omega^{(u,v)}_i}|}\sum_{P_j \in {\mathit{\Omega^{(u,v)}_i}}}\mathit{D_{KL}}(P^{(u,v)}_i ||  P^{(u,v)}_j),
\end{equation}

where $D_{KL}(P^{(u,v)}_i || P^{(u,v)}_j)$ is the KL Divergence between distributions $P^{(u,v)}_i$ and $P^{(u,v)}_j$. A high view divergence score implies that on average, the prediction in the current view is significantly different to the predictions coming from other views (Fig.~\ref{fig:score_vis}).

\subsection{Region Selection}
\label{sec:region_selection}

To exploit the structure and locality in semantic segmentation masks, we argue for selecting regions to be labeled instead of whole images. In particular, we opt for using superpixels since they are most of the time associated with a single object-class and therefore lightweight to label for the annotator. Fixed sized rectangular regions would have been another option, but most of the time contain more than one object-class, leading to more effort for the annotator to precisely delineate the boundary between objects. Our implementation uses the SEEDS~\cite{van2012seeds} algorithm for superpixel computation.

For each superpixel $r$, the two scores $\mathit{VE}_{i}^r$ and $\mathit{VD}_{i}^r$ are computed as average of the view entropy and view divergence of all the pixels in $r$
\begin{equation}
    \mathit{VE_{i}^{r}} = \frac{1}{|r|}\sum_{(u,v) \in r}\mathit{VE}_{i}^{(u,v)}
\end{equation}
\begin{equation}
    \mathit{VD_{i}^{r}} = \frac{1}{|r|}\sum_{(u,v) \in r}\mathit{VD}_{i}^{(u,v)},
\end{equation}
with $|r|$ is the number of pixels in superpixel $r$.

Our strategy to select the next superpixel to label consists of two steps. First, we look for the superpixel $r$ from image $I_i$ that has the highest view entropy:
\begin{equation}
    (i, r) = \argmax_{(j,s)} \mathit{VE^{s}_j}
\end{equation}
 
Then, we identify the set of superpixels in the dataset whose cross-projection overlap is at least with 50\% of $(i, r)$, including self, and denote this set as $\mathit{S}$. We then look for the superpixel from $\mathit{S}$ that has the highest view divergence as:
\begin{equation}
    (k, t) = \argmax_{(j, s) \in \mathit{S}} \{ \mathit{VD^{s}_j \mid (j, s)\text{ and }(i, r)\text{ \textit{overlap} }} \}
\end{equation}

All the superpixels in $\mathit{S}$ are then removed from further selection considerations. This selection process is repeated until we select superpixels equivalent to the requested $K$ images.

\subsection{Label Acquisition}

\begin{figure}[t]
    \begin{center}
   \includegraphics[width=\linewidth]{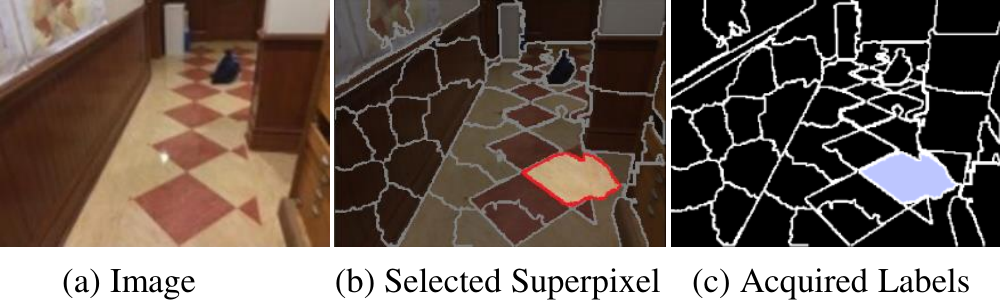}
    \end{center}
    \vspace{-0.4cm}
	\caption{Label acquisition. We ask the oracle to label only the superpixels selected by our method (marked red in (b)). The remaining superpixels of the ground-truth map, shown in black in (c), are marked with the \textit{ignore} label.}\label{fig:label_acquisition}
	
\end{figure}

Next, we acquire labels for superpixels selected in Section \ref{sec:region_selection}. Instead of using a real annotator, we simulate annotation by using the ground truth annotation of the superpixels as the annotation from the oracle. These labeled regions are then added to the labeled dataset and removed from the unlabeled dataset. The labeled dataset therefore is comprised of a lot of images, each with a subset of their superpixels labeled. The unlabeled superpixels of these images are marked with the \textit{ignore} label (Fig. \ref{fig:label_acquisition}).

The active selection iteration concludes with the retraining of the network with the updated dataset $D_L$ from scratch. 

% !TEX root = main.tex

\section{Results}

\subsection{Datasets and Experimental Settings}
We evaluate our approach on three public datasets, SceneNet-RGBD~\cite{McCormac:etal:ICCV2017,McCormac:etal:arXiv2016}, ScanNet~\cite{dai2017scannet}, and Matterport3D~\cite{chang2017matterport3d}. 
All datasets have a large number of RGBD frames across indoor scenes along with their semantic annotations. 
SceneNet-RGBD is a synthetic dataset containing large-scale photorealistic renderings of indoor scene trajectories. 
ScanNet contains around 2.5M views in 1513 real indoor scenes. 
Matterport3D has around 200K RGBD views for 90 real building-scale scenes.
We use a subset of images (72290, 23750, and 25761 for SceneNet-RGBD, ScanNet, and Matterport3D, respectively), since active learning iterations on the entire datasets would be too expensive in terms of compute. 
Further, we resize all images to a resolution of $320\times240$ pixels. 
We refer to the supplementary material for statistics of dataset subsets.
We use the official train/test scene splits for training and evaluation for ScanNet and Matterport3D. 
For SceneNet-RGBD, as train set, we use every tenth frame from 5 of the 17 training splits, and for validation, every 10th frame from the validation split. 
The seed set is initialized with fully annotated 1.5\% of training images that are randomly selected from the unlabeled dataset. 
For ScanNet and Matterport3D, the mIoU is reported on the test set on convergence. For SceneNet, we report the mIoU on the validation set, since a public test set is not available. The model is considered converged when the mIoU of the model does not increase within 20 epochs on the validation set. For ScanNet and Matterport3D the networks are trained with SGD with an initial learning rate of 0.01, while for SceneNet-RGBD we use a learning rate of 0.005. The learning rate is decayed on the 40th epoch to 0.1 times its original value. Further, we set momentum to 0.9, and weight decay penalty to 0.0005.

\subsection{Comparisons against Active Learning Methods}
We compare our method against 9 other active selection strategies. 
These include random selection (\textbf{RAND}), softmax margin (\textbf{MAR})~\cite{joshi2009multi, roth2006margin, wang2016cost}, softmax confidence~\cite{settles2008analysis, wang2016cost} (\textbf{CONF}), softmax entropy~\cite{hwa2004sample, wang2016cost} (\textbf{ENT}), MC dropout entropy~\cite{gal2017deep} (\textbf{MCDR}), Core-set selection~\cite{sener2017active} (\textbf{CSET}), maximum representativeness~\cite{yang2017suggestive} (\textbf{MREP}), CEAL entropy~\cite{wang2016cost} (\textbf{CEAL}), and regional MC dropout entropy~\cite{mackowiak2018cereals} (\textbf{RMCDR}).
For all methods, we use the same segmentation network (DeepLabv3+~\cite{chen2018encoder}); we also use the same random seed during sample selections for a fair comparison. 
At the end of each active iteration, the active selection algorithm chooses the next $K$ samples to be labeled ($K=1500, 1250$, and $1000$ for SceneNet-RGBD, ScanNet and Matterport3D, respectively). 
We use 40 superpixels per image in our method and a window size of $40 \times 40$ in the case of \textbf{RMCDR}. 
For further details, refer to the supplementary material.

\begin{figure*}
  \includegraphics[width=\textwidth]{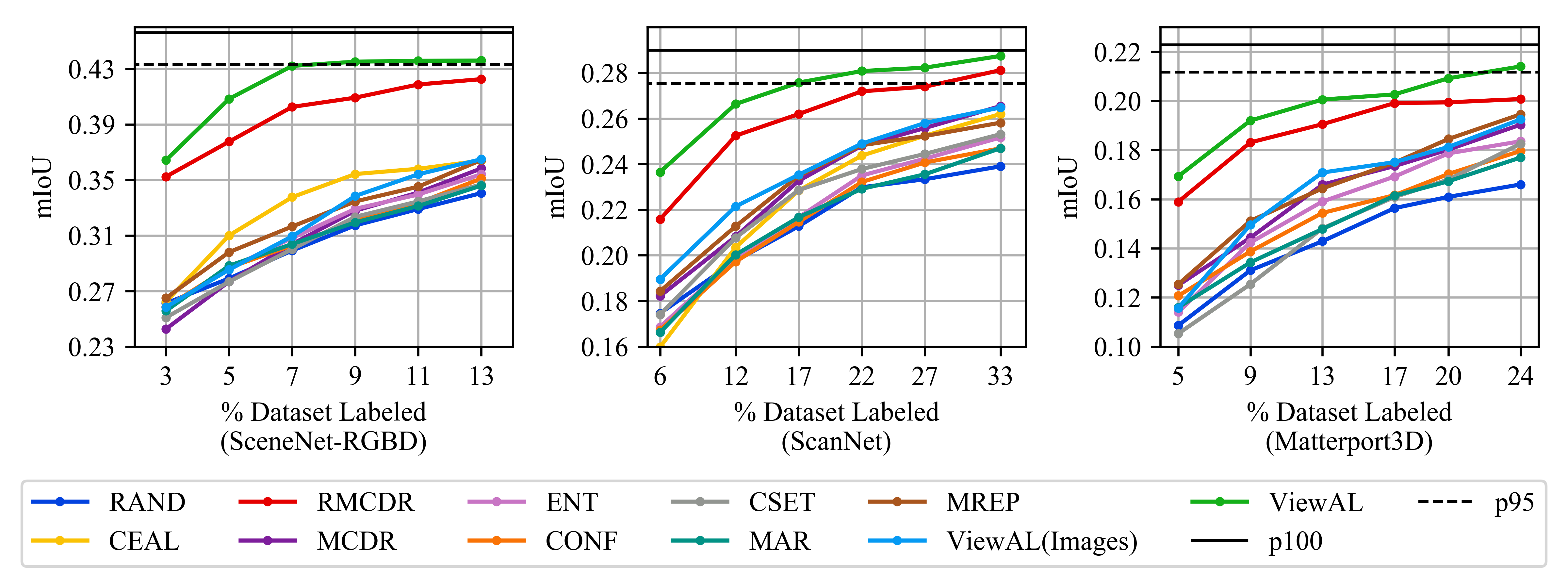}
  \vspace{-5mm}
  \caption{Active learning performance for our method and other baselines. The horizontal solid line (p100) at the top represents model performance with the entire training set labeled. The dashed line (p95) represents 95\% of that performance. We observe that our method outperforms all other methods and is able to achieve 95\% of maximum model performance with just 7\%, 17\% and 24\% labeled data on SceneNet-RGBD~\cite{McCormac:etal:arXiv2016}, ScanNet~\cite{dai2017scannet}, and Matterport3D~\cite{chang2017matterport3d} datasets. Note that we omit the results with the seed set here since all methods have the same performance on it.}\label{fig:results}
  \vspace{-5mm}
\end{figure*}

Fig.~\ref{fig:results} shows the mIoU achieved by the segmentation model against percentage of dataset labeled used to train the model. 
All the active learning methods outperform random sampling (\textbf{RAND}).
Our method achieves a better performance than all other methods for the same percentage of labeled data. 
On ScanNet, with just 17\% labeled data, we are able to obtain an mIoU of 27.7\%.
This is around 95\% of the model performance achieved when trained over the whole dataset (28.9\%) on ScanNet dataset. 
The method that comes closest to ours (\textbf{RMCDR}) achieves the same performance with 27\% data. 
We outperform other methods on SceneNet-RGBD and Matterport3D dataset as well, where we achieve 95\% of maximum model performance with just 7\% and 24\% labeled data respectively, while the runner-up method does so with 14\% and 33\% data on the respective datasets.
The results are presented in tabular form in the supplementary material.

\subsection{Evaluation on Labeling Effort}

The previous experiment compared active learning performance as a function of proportion labeled data used for training. However, it gives no indication of the effort involved in labeling that data. 
In order to give a clear indication of the actual effort involved, we compare the time taken to annotate images to the time taken to annotate superpixels. 

\begin{figure}[h]
\begin{center}
   \includegraphics[width=0.99\linewidth]{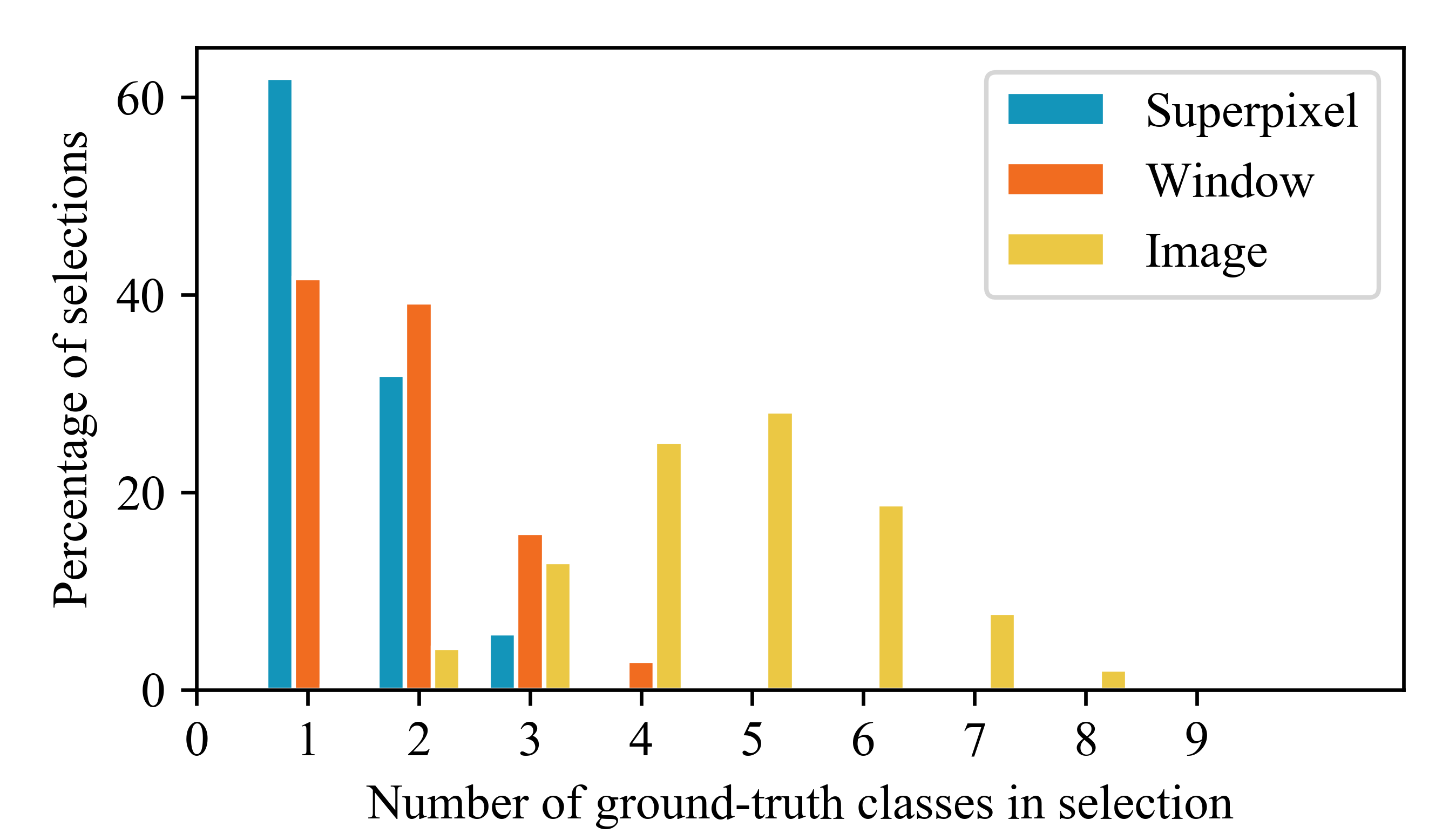}
\end{center}
\vspace{-0.4cm}
   \caption{Distribution of number of unique ground truth classes in selections made by different approaches. The majority of the selections made by our superpixel approach have only a single ground truth label in them. Compared to the fixed window approach which has an expected 1.83 unique classes per selection, our approach has 1.40 expected classes per selection. Here, the average number of pixels per window and per superpixel was taken to be the same.}
\label{fig:distribution_labels}
\vspace{-0.2cm}
\end{figure}
Since the majority of selections made by our approach have just a single ground truth label associated with them (Fig. \ref{fig:distribution_labels}), it is expected that they will require less labeling time.
We verified this in a user study, where we asked users to label 50 images and their equivalent 2000 superpixels ($40$ superpixels per image) selected from the ScanNet dataset. 
The images were labeled using a LabelMe~\cite{russell2008labelme}-like interface, LabelBox\footnote{https://labelbox.com/}, while superpixels were labeled using a superpixel-based interface, which allowed boundary corrections. 
We made sure that the quality of labels for both images and superpixels was comparable (0.953 mIoU) and acceptable (0.926 and 0.916 mIoU against ground-truth, respectively).
Further, to reduce the variance in time due to user proficiency, we made sure that an image and its respective superpixels were labeled by the same user.

\begin{figure}[h]
\begin{center}
   \includegraphics[width=\linewidth]{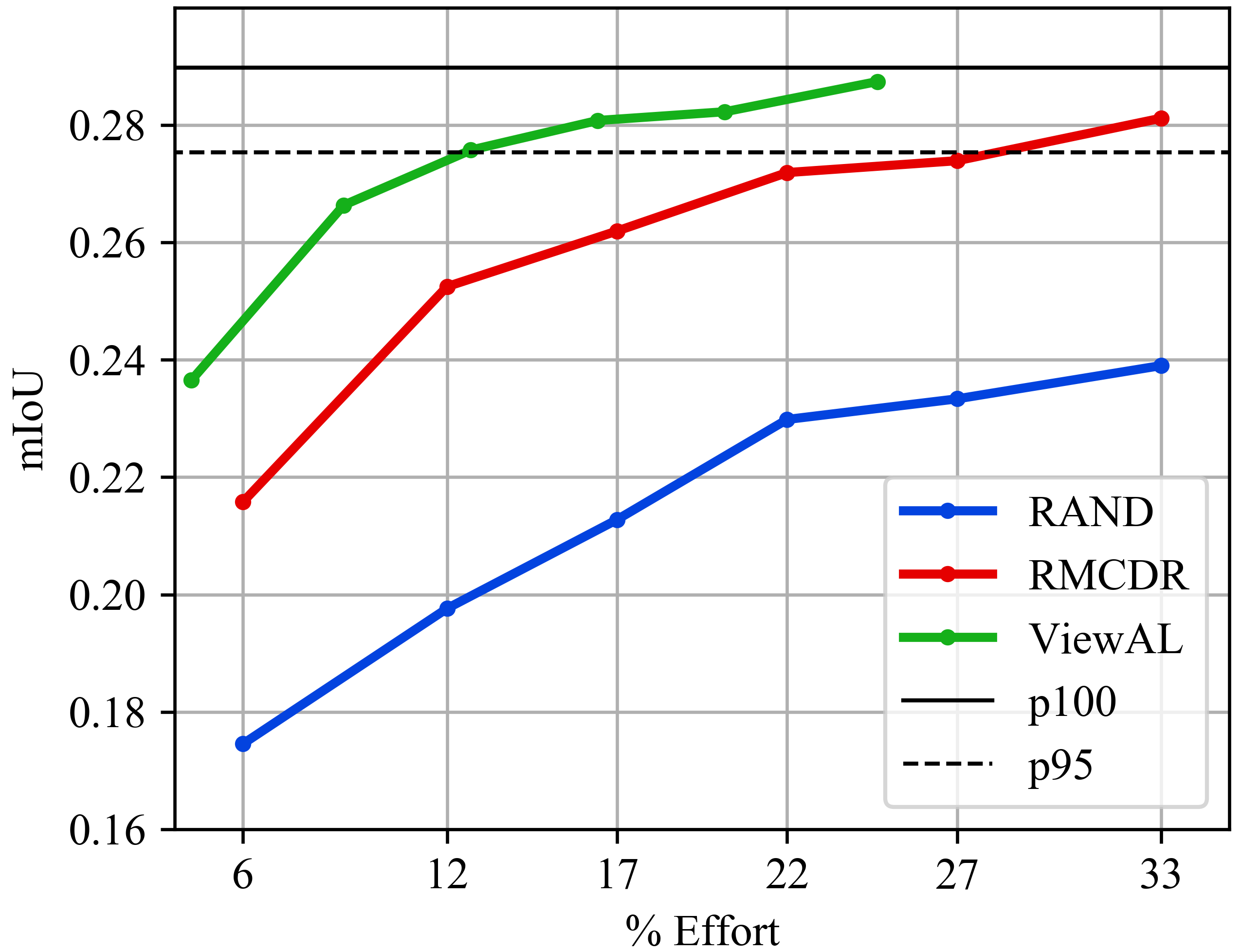}
\end{center}
\vspace{-0.4cm}
   \caption{mIoU vs Labeling Effort on the ScanNet dataset. One unit of effort is the expected time taken to label 1\% of the dataset using a LabelMe~\cite{russell2008labelme}-like interface. Our method delivers 95\% of maximum model performance while requiring only 13\% effort.}
\label{fig:effort}
\end{figure}

% Results
It took our annotators 271 minutes to annotate 50 images, while it took them just 202 minutes to annotate their superpixels, therefore bringing down the time by 25\% and demonstrating that using superpixels indeed reduces labeling effort. 
Fig. \ref{fig:effort} compares performance of our active learning as a function of labeling effort involved, taking into account the reduction in effort due to superpixel-wise labeling. Effort here is measured in terms of time, with 1\% effort being equivalent to the time taken to annotation 1\% of data using a LabelMe-like interface. Our method achieves 95\% maximum model performance with just 13\% effort, compared to 27\% effort by the runner-up method on the ScanNet dataset.

\subsection{Ablation Studies}
\begin{figure}[t]
\begin{center}
   \includegraphics[width=\linewidth]{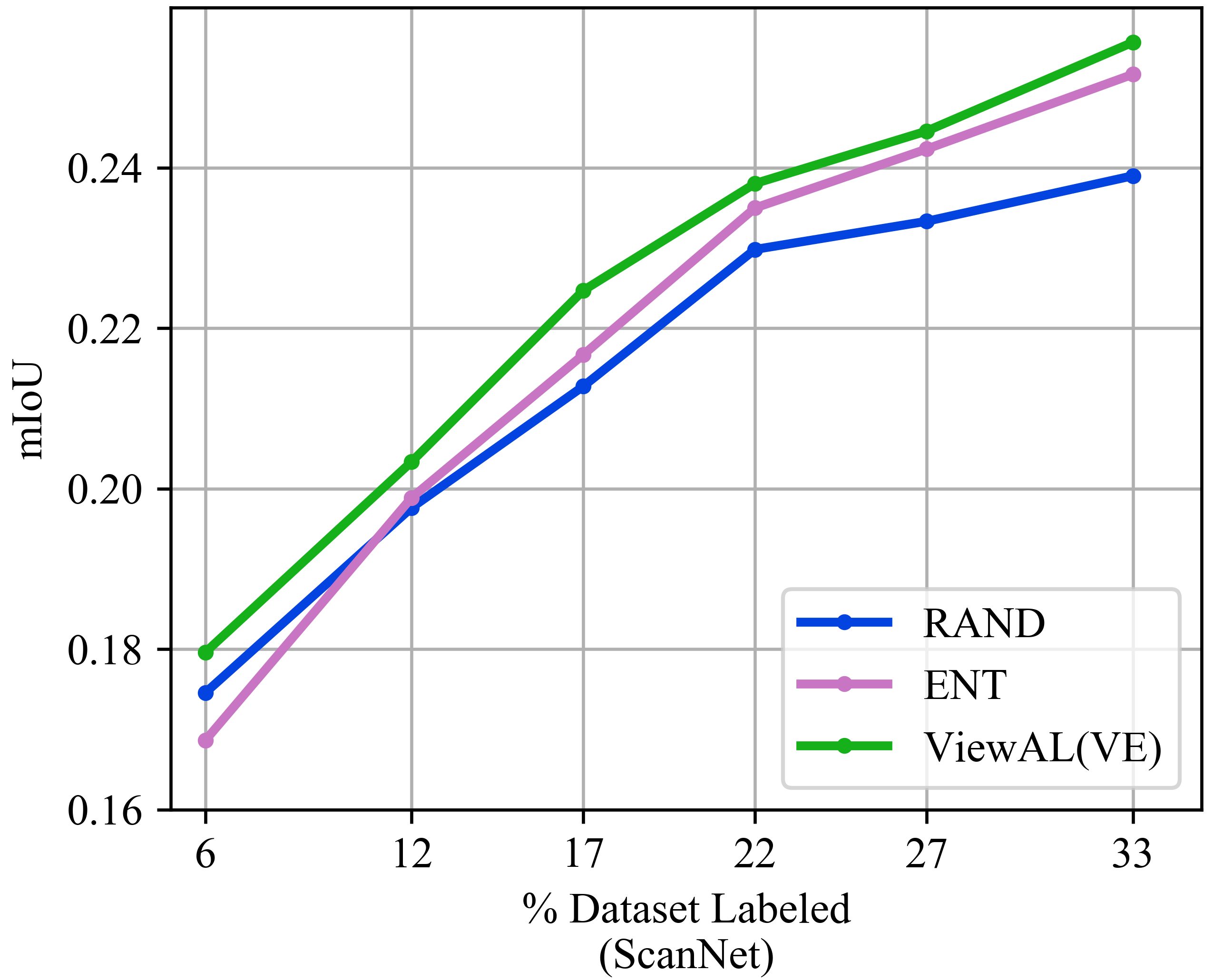}
\end{center}
\vspace{-0.4cm}
   \caption{Results using only view entropy (i.e., without superpixels, MC dropout, or view divergence) which outperforms both random (RAND) and softmax entropy (ENT) methods.}
\label{fig:view_only_scannet}
\end{figure}

\paragraph{Effect of view entropy in isolation.} 
In order to show that view entropy provides a helpful signal for active selection, we evaluate the performance of a non-regional variant (i.e., no superpixel selection) without the use of MC dropout or view divergence.
In this case, $P^{(u,v)}_i(c)$ from Eq.~\ref{eqn:mcdropout_p} is just the softmax probability of the network (instead of average across MC dropout runs).
That means that images are only selected on the basis of the view entropy score (Section \ref{section:view_entropy}), which is averaged across all pixels of the image.
Fig.~\ref{fig:view_only_scannet} shows that view entropy in isolation significantly helps selection while outperforming both \textbf{RAND} and \textbf{ENT} methods.

\begin{figure}[t]
\begin{center}
   \includegraphics[width=\linewidth]{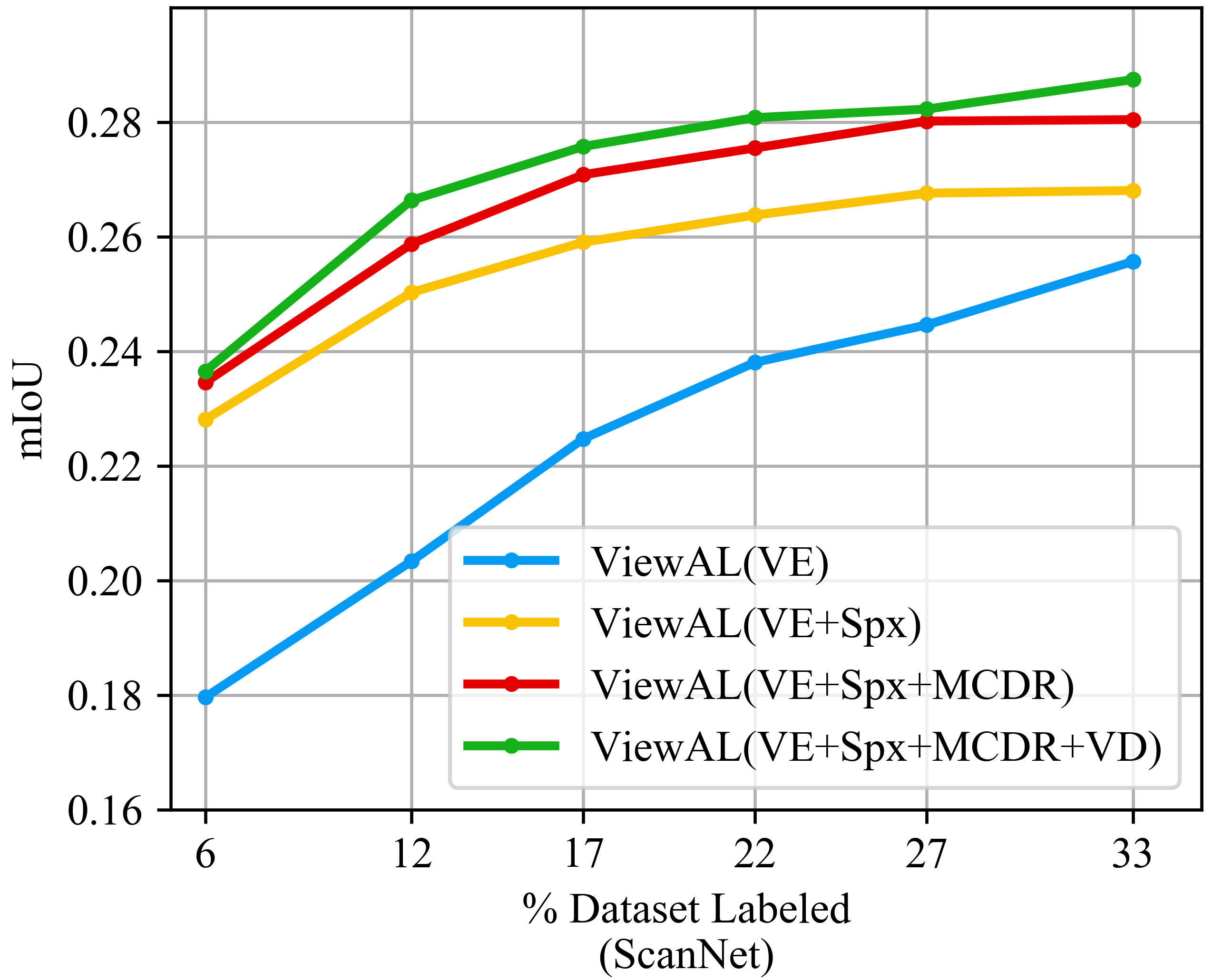}
\end{center}
\vspace{-0.4cm}
   \caption{Ablation study of our method: \OURS(VE) is our method without superpixels, MC dropout, and view divergence. When superpixels are used for selection over entire images, we see significant improvements as shown by the curve \OURS(VE+Spx). Adding MC dropout improves performance further as indicated by \OURS(VE+Spx+MCDR). Our final method, \OURS(VE+Spx+MCDR+VD) improves over this further by adding view divergence.}
\label{fig:ablation_scannet}
\end{figure}

\vspace{-0.2cm}
\paragraph{Effect of superpixels.} 
We evaluate the effect of using superpixel selection in combination with view entropy scores. 
In Fig.~\ref{fig:ablation_scannet}, the curve \OURS(VE+Spx) shows the effectiveness of selecting superpixels rather than entire images. 
Active learning performance significantly improves as superpixel-based region selection facilitates focus only on high scoring regions while ignoring less relevant ones.

\paragraph{Effect of using MC dropout.} 
Instead of using the plain softmax probabilities as in the last two paragraphs, we use MC dropout to get an estimate of class probabilities as shown in Eq.~\ref{eqn:mcdropout_p}. 
The curve \OURS(VE+Spx+MCDR) in Fig.~\ref{fig:ablation_scannet} shows that using MC dropout improves active learning performance by a considerable margin. 
This can be explained by MC dropout providing a better estimate of class posteriors than just using simple softmax probabilities.
\paragraph{Effect of using view divergence.} 
Finally, we add view divergence score (Section \ref{section:view_divergence}), giving us our complete method. 
View divergence helps select the view for a superpixel with the most different class-wise probability distribution from the rest of views that superpixel appears in. 
It further improves the active learning performance of our method as shown by the curve \OURS(VE+Spx+MCDR+VD) in Fig. \ref{fig:ablation_scannet}.
\newline

% !TEX root = main.tex

\section{Conclusions}
We have introduced \OURS{}
%\footnote{Source code available: \href{https://github.com/nihalsid/ViewAL}{https://github.com/nihalsid/ViewAL}}
, a novel active learning method for semantic segmentation that leverages inconsistencies in prediction across views as a measure of uncertainty. 
In addition to a new view entropy score, we propose to use regional selection in the form of superpixels, which we incorporate in a unified active learning strategy.
This allows us to acquire labels for only the most promising areas of the images and at the same time reduce the labeling effort. 
While we have shown results on widely-used indoor datasets, our method is agnostic to the respective multi-view representation, and can easily be applied to a wide range of scenarios.
%
%In our experiments, cross projections used for making associations between the pixels required depths and poses. 
%
%These can alternatively be obtained by using structure from motion/multi-view stereo methods like {COLMAP} \cite{schonberger2016pixelwise,seitz2006comparison}. 
%
Currently, our focus is on semantic segmentation; however, we believe that this work provides a highly promising research avenue towards other tasks in computer vision, including instance segmentation, object detection, activity understanding, or even visual-language embeddings.

\section*{Acknowledgements}
This work is funded by Google (Daydream - AugmentedPerception), the ERC Starting Grant Scan2CAD (804724), and a Google Faculty Award.
We would also like to thank the support of the TUM-IAS, funded by the German Excellence Initiative and the European Union Seventh Framework Programme under grant agreement n° 291763, for the TUM-IAS Rudolf M{\"o}{\ss}bauer Fellowship.
Finally, we thank Angela Dai for the video voice over.

{\small
\bibliographystyle{ieee_fullname}
\bibliography{egbib}
}
\vfill\eject
\appendix
\section{Dataset Statistics}
\label{app:stats}
We evaluate our approach on three public datasets, SceneNet-RGBD~\cite{McCormac:etal:ICCV2017,McCormac:etal:arXiv2016}, ScanNet~\cite{dai2017scannet}, and Matterport3D~\cite{chang2017matterport3d}. 
SceneNet-RGBD is a synthetic dataset containing large-scale photorealistic renderings of indoor scene trajectories, with around 5M RGBD frames. 
ScanNet contains around 2.5M views in 1513 real indoor scenes. 
Matterport3D has around 200K RGBD views for 90 real building-scale scenes.
We use a subset of images, as highlighted in Table~\ref{table:dataset_stats}, since active learning iterations on the entire datasets would be too expensive in terms of compute. 
\begin{table}[h]
\begin{center}
\begin{tabular}{|l|l|l|l|}
\hline
Statistic & \pbox{5cm}{SceneNet}  & ScanNet & Matterport3D \\
\hline\hline
Train Sequences & 2434 & 1041 & 968\\
\hline
Train Frames & 72990 & 23750 & 25761 \\
\hline
Validation Seqs. & 500 & 465 &  214\\
\hline
Validation Frames & 15000 & 5453 & 13702\\
\hline
Test Sequences & - & 80 & 370\\
\hline
Test Frames & - & 5320 & 22588\\
\hline
Semantic Classes & 13 & 40 & 40\\
\hline
\end{tabular}
\end{center}
\caption{Statistics of SceneNet-RGBD\cite{McCormac:etal:arXiv2016}, ScanNet\cite{dai2017scannet} and Matterport3D\cite{chang2017matterport3d} dataset subsets used in our experiments. }
\label{table:dataset_stats}
\end{table}

\section{Baseline Active Learning Methods}
\label{app:methods}
We compare our method against popular uncertainty and diversity based active learning approaches found in the literature. Here, we give a brief overview of these approaches. In terms of notation,  $D_U$ is the unlabeled dataset, $D_L$ is the currently labeled dataset, $M$ is the total number of target classes, $K$ is the number of images from $D_U$ requested to be labeled in each active selection iteration, $n$ goes over pixels for image $i$ and $\theta_{SEG}$ are the parameters of the segmentation network. 
\paragraph{Random Selection (RAND)}
In random selection, in each active selection iteration, the next query for $K$ samples is composed of randomly selected samples from the unlabeled dataset.
\paragraph{Softmax Confidence (CONF)}
The least confidence approach discussed in \cite{settles2009active} can be adapted to deep convolutional networks by using softmax probability of the most probable class as confidence~\cite{wang2016cost}. This selection strategy then selects the least $K$ confident samples from $D_U$ as the next query. For semantic segmentation, we calculate confidence for each pixel and use the sum across pixels as the confidence for the image. For each image $i$, the confidence score is therefore given by Eq.~\ref{eqn:conf}, and $K$ least scoring samples are selected for label acquisition.
\begin{equation} \label{eqn:conf}
    S_{i}^{CONF} = \sum_{n}{\max\limits_{j}}\:p(y_{i}^{n} = j\:|\:x_i;\theta_{SEG})
\end{equation}
\paragraph{Softmax Margin (MAR)}
Similar to CONF, this approach \cite{scheffer2001active} ranks all the samples in order of the difference of softmax probabilities of the most probable label ($j_1$) and the second most probable label ($j_2$), and chooses the $K$ samples which have the least difference (Eq.~\ref{eqn:mar})~\cite{wang2016cost}. The idea is that samples for which the network has a small margin between the top predictions means that the network is very uncertain between the two. 
\begin{equation} \label{eqn:mar}
    S_{i}^{MAR} = \sum_{n}(p(y_{i}^{n} = j_1\:|\:x_i;\theta_{SEG}) - p(y_{i}^{n} = j_2\:|\:x_i;\theta_{SEG}))
\end{equation}
\paragraph{Softmax Entropy (ENT)} In the case of semantic segmentation, the entropy value for each pixel in the image is summed to get the entropy score for the whole image (Eq.~\ref{eqn:ent}). 
\begin{equation}\label{eqn:ent}
    S_{i}^{ENT} = -\sum_{n}\sum_{j=1}^{M}p(y_{i}^{n}\:|\:x_i;\theta_{SEG})\log{p(y_{i}^{n}\:|\:x_i;\theta_{SEG})}
\end{equation}
Entropy takes into account probabilities of all classes unlike CONF, which considers most probable class or MAR, which only considers the top two most probable classes.
\paragraph{CEAL Entropy (CEAL)}
CEAL \cite{wang2016cost} combines CONF, MAR, ENT methods with pseudo-labeling in their active learning framework. We only compare with their ENT variant since the results are quite identical for all the other measures. At the end of each active selection iteration, they propose not only adding samples labeled by the oracle, but also high confidence samples from $D_U$ for which softmax entropy is less than the threshold $\delta$. For these samples, the assigned labels are the predicted ones by the current model. The idea behind pseudo-labeling is that since the high confidence samples are close to the labeled samples in CNNs feature space, adding them in training is a reasonable data augmentation for CNN to learn robust features. Further, as the active iteration increase, the number of samples selected for pseudo-labeling increases since the network gets more and more confident. To prevent high amounts of pseudo-labeling, the threshold is decreased at the end of each selection iteration. Our implementation of CEAL only assigns pseudo-labels at pixel level instead of image level to account for locality of segmentation task. 
\paragraph{Monte Carlo Dropout (MCDR)}
It has been argued in \cite{gal2016dropout} that vanilla deep learning models rarely represent model uncertainty, and softmax entropy is not really a good measure of uncertainty. Instead of softmax probabilities~\cite{gal2017deep} use Monte Carlo (MC) dropout to estimate model uncertainty. 
\begin{equation}
\label{eqn:mcdropout_s}
\resizebox{0.5\textwidth}{!}{
        $S_{i}^{MCDR} = -\sum_{n}\sum_{j=1}^{M}p_{MC}(y_{i}^{n}\:|\:x_i;\theta_{SEG})\log{p_{MC}(y_{i}^{n}\:|\:x_i;\theta_{SEG})}$
    
}
\end{equation}
where $p_{MC}$ is given by
\begin{equation}
\label{eqn:mcdropout_p}
        p_{MC}(y_{i}^{n}\:|\:x_i;\theta_{SEG}) = \frac{1}{D}\sum_{d=1}^{D}{p_{SM}(y_{i}^{n}\:|\:x_i;\theta^d_{SEG})},
\end{equation}
with $D$ being the total number of MC Dropout runs.
\paragraph{Regional MC Dropout (RMCDR)}
Proposed for semantic segmentation in \cite{mackowiak2018cereals}, it follows the same approach as MCDR. However, instead of calculating scores for whole images, scores are calculated for fixed-size regions. The selection algorithm is then selecting as many highest entropy regions as it takes to make up $K$ images. The original method of~\cite{mackowiak2018cereals} uses Vote Entropy~\cite{dagan1995committee}, however we use MC Dropout since it gives slightly better results. Further, the method of~\cite{mackowiak2018cereals} uses cost estimates regressed from annotator click patterns, which we dont use since these are not available for any of the datasets we evaluate on.    
\paragraph{Maximum Representativeness (MREP)}
Unlike the other approaches discussed until now which were only uncertainty based, MREP is a mixed approach that combines uncertainty and diversity. This method, proposed in \cite{yang2017suggestive} first choose points that are highly uncertain. From among these points, it further chooses points that best represent the rest of distribution based on some similarity measure. In our implementation, vote entropy is first used to select $2K$ samples, and then $K$ most representative samples amongst those are selected to be labeled. We use the Euclidean norm for the similarity measure.
\paragraph{Core-Set Selection (CSET)}
Core-Set \cite{Sener2017} is a purely diversity-based approach. The method aims to select a subset of $K$ points such that the model trained on a subset of the points is competitive for the rest of the points. The $K$ samples selected are the ones that have the smallest $\delta$ for the $\delta$ cover of the set. This means that the algorithm seeks to minimize the maximum distance between sample $x_i$ in the remaining unlabeled dataset and its closest neighbor $x_j$ in the selected subset. We use the simple greedy selection strategy proposed in \cite{Sener2017} as it performs only slightly worse than the robust version.

\section{Performance with Imperfect Depth and Pose}
Here we evaluate the performance of our method when the ground truth depth and pose are not available, i.e. only RGB frames are available. In such a case, one alternative for making associations between pixels across frames is to use structure from motion/multi-view stereo methods. We use {COLMAP} \cite{schonberger2016pixelwise,seitz2006comparison} to first reconstruct the scenes from RGB frames and obtain depth and camera parameters. We use 5 scenes from ScanNet~\cite{dai2017scannet} for this. We keep to just 5 scenes as the time taken to reconstruct a scene using {COLMAP} is quite long, and since here we only want to compare the performance using ground truth depth and pose against reconstructions, these should be sufficient. We use 1000 frames from each scene, and split the total 5000 frames into 2000 training (unlabeled), 1000 validation and 2000 test frames. The seed set has 100 fully labeled frames. Each selection iteration chooses 100 more frames (or equivalent superpixels) from the training set to be labeled. We compare against random selection (\textbf{RAND}) and the variant of our method that uses true pose and depth ({ViewAL(TRUE)}).

\begin{figure}[t]
\begin{center}
   \includegraphics[width=\linewidth]{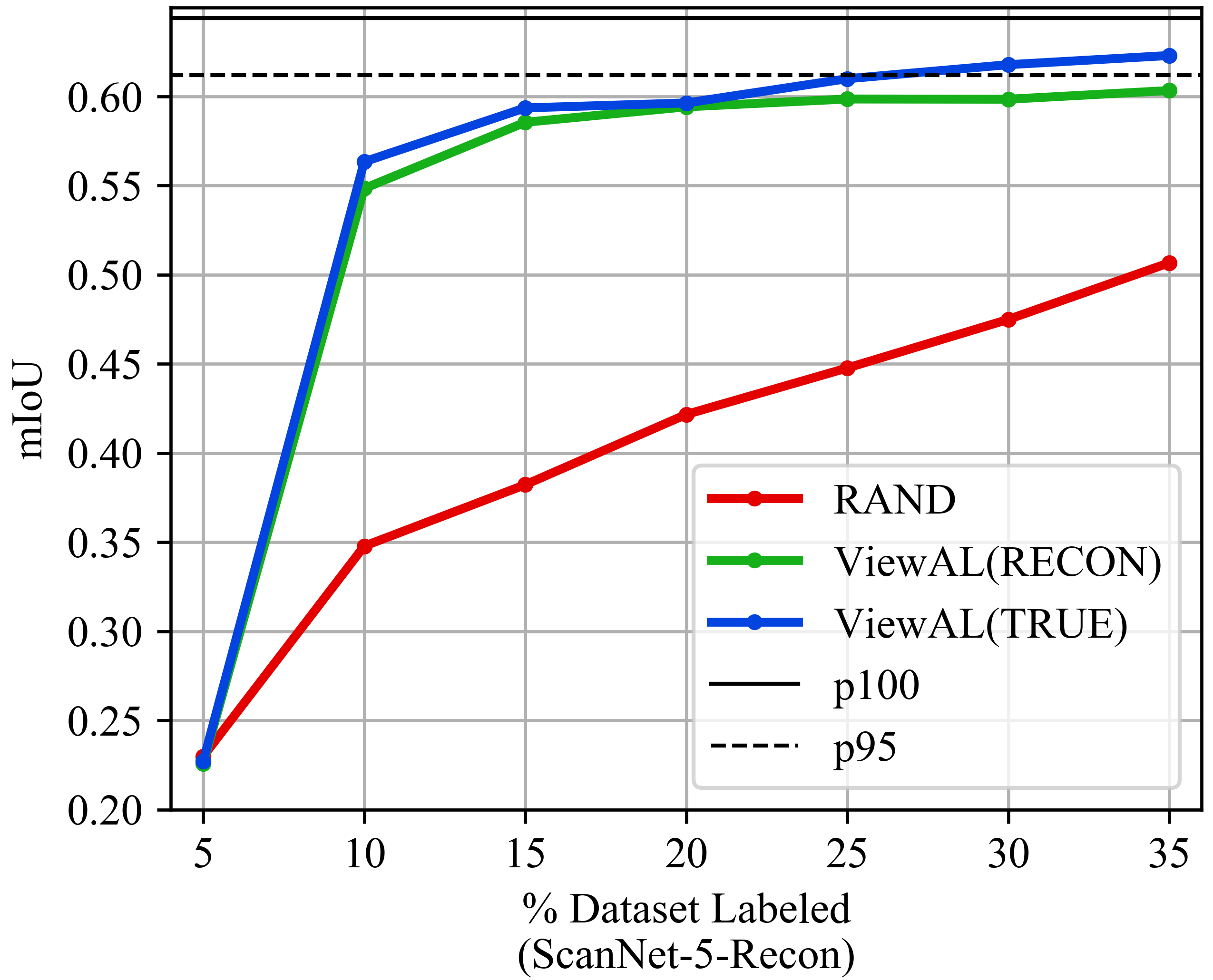}
\end{center}

   \caption{Performance with imperfect depth and pose. Our method using reconstructed depth and pose, ViewAL(RECON), outperforms the RAND baseline and performs only slightly worse than the variant using true depth and poses, ViewAL(TRUE).}
\label{fig:colmap}
\end{figure}

Fig. \ref{fig:colmap} shows the results for this experiment. We observe that our method which uses reconstructed depth and pose still outperforms the \textbf{RAND} baseline and performs only slightly worse than the variant using true depth and poses.

\section{Comparison with baselines allowed to select superpixels}
\begin{figure}[t]
\begin{center}
   \includegraphics[width=\linewidth]{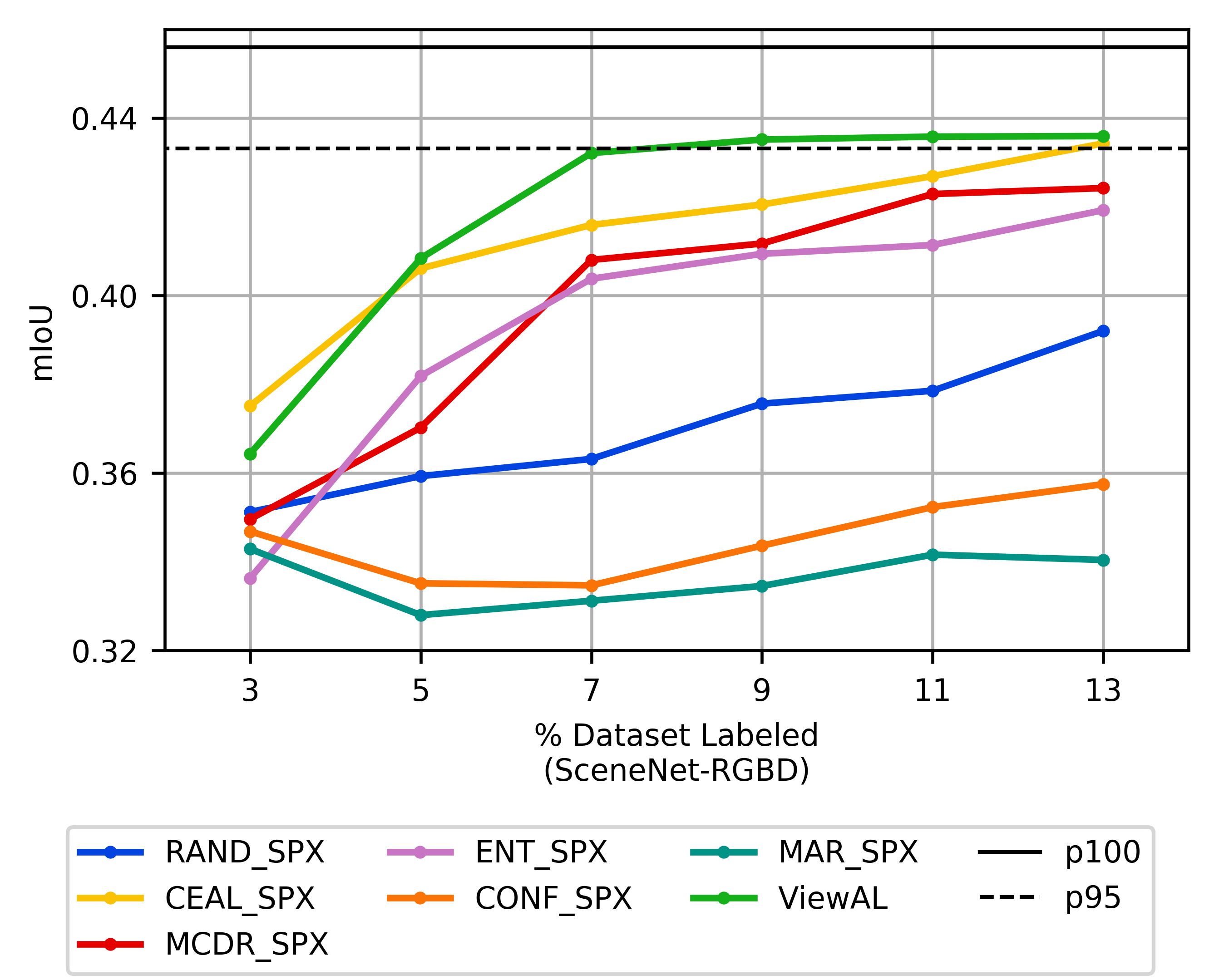}
\end{center}

   \caption{Active learning performance for our method and other baselines when all baselines use superpixels.}
\label{fig:albase}
\end{figure}

Fig.~\ref{fig:albase} shows the scenario where other methods are allowed to use superpixel selection instead of window / image selection. It can be observed that most methods do benefit from superpixel selection.

\section{Handling non-static data}
For computation of view entropy and divergence scores, we need to associate superpixels between frames. 
In our experiments, we use frame depth and pose to get these associations.
However, this can be done only in case of static scenes, i.e. when objects do not change positions across frames.
A promising future direction could be to extend this work for the dynamic setting using, for instance, optical flow estimates or keypoint descriptor matching to achieve superpixel association across frames.
\section{Result Tables}
\label{app:tables}
Due to limited space in the main paper, we present the experimental results here in tabular form. Table~\ref{tab:res_scenenet}, Table~\ref{tab:res_scannet}, Table~\ref{tab:res_matterport} list results for all the methods we compared on SceneNet-RGBD~\cite{McCormac:etal:arXiv2016}, ScanNet~\cite{dai2017scannet} and Matterport3D~\cite{chang2017matterport3d} datasets. Table~\ref{tab:ablation} reports results for the ablation study.
\begin{table*}[t]
  \centering
  \small
  \begin{tabular}{|l|l|l|l|l|l|l|l|l|l|l|l|}
  \hline
\pbox{15cm}{\% Labeled \\ Data}&RAND&RMCDR&MCDR&ENT&CONF&CSET&MAR&MREP&CEAL&\pbox{15cm}{ViewAL\\(Images)}&ViewAL\\
\hline\hline
1 & 0.2245 & 0.2124 & 0.2160 & 0.2261 & 0.2257 & 0.2259 & 0.2254 & 0.2168 & 0.2255 & 0.2159 & 0.2125\\\hline
3 & 0.2612 & 0.3524 & 0.2427 & 0.2586 & 0.2584 & 0.2509 & 0.2558 & 0.2650 & 0.2624 & 0.2585 & \textbf{0.3643}\\\hline
5 & 0.2791 & 0.3776 & 0.2768 & 0.2864 & 0.2868 & 0.2767 & 0.2882 & 0.2980 & 0.3101 & 0.2854 & \textbf{0.4084}\\\hline
7 & 0.2991 & 0.4026 & 0.3038 & 0.3082 & 0.3029 & 0.3001 & 0.3038 & 0.3165 & 0.3376 & 0.3094 & \textbf{0.4321}\\\hline
9 & 0.3173 & 0.4092 & 0.3278 & 0.3292 & 0.3208 & 0.3234 & 0.3194 & 0.3345 & 0.3542 & 0.3385 & \textbf{0.4352}\\\hline
11 & 0.3290 & 0.4187 & 0.3409 & 0.3395 & 0.3334 & 0.3346 & 0.3313 & 0.3451 & 0.3580 & 0.3541 & \textbf{0.4358}\\\hline
13 & 0.3405 & 0.4226 & 0.3583 & 0.3541 & 0.3510 & 0.3467 & 0.3459 & 0.3644 & 0.3639 & 0.3649 & \textbf{0.4359}\\\hline
15 & 0.3509 & 0.4337 & 0.3716 & 0.3616 & 0.3630 & 0.3285 & 0.3522 & 0.3755 & 0.3781 & -  & \textbf{0.4383}\\\hline
17 & 0.3587 & 0.4340 & 0.3737 & 0.3726 & 0.3731 & 0.3432 & 0.3688 & 0.3845 & 0.3807 & -  & \textbf{0.4412}\\
\hline
  \end{tabular}
  \vspace{0.3cm}
  \caption{Semantic segmentation performance in terms of mIoU when labeled data is selected using baseline active learning methods and our method on SceneNet-RGBD~\cite{McCormac:etal:arXiv2016} dataset.}
  \label{tab:res_scenenet}
\end{table*}

\begin{table*}[h!]
  \centering
  \small
  \begin{tabular}{|l|l|l|l|l|l|l|l|l|l|l|l|}
  \hline
\pbox{15cm}{\% Labeled \\ Data}&RAND&RMCDR&MCDR&ENT&CONF&CSET&MAR&MREP&CEAL&\pbox{15cm}{ViewAL\\(Images)}&ViewAL\\
\hline\hline
1 & 0.0998 & 0.0957 & 0.0950 & 0.0961 & 0.0958 & 0.0961 & 0.0999 & 0.0934 & 0.1001 & 0.0957 & 0.0953\\\hline
6 & 0.1746 & 0.2158 & 0.1821 & 0.1686 & 0.1672 & 0.1741 & 0.1662 & 0.1843 & 0.1598 & 0.1895 & \textbf{0.2365}\\\hline
12 & 0.1976 & 0.2525 & 0.2083 & 0.1989 & 0.1972 & 0.2077 & 0.2003 & 0.2128 & 0.2035 & 0.2214 &\textbf{0.2663}\\\hline
17 & 0.2128 & 0.2619 & 0.2327 & 0.2167 & 0.2146 & 0.2286 & 0.2167 & 0.2349 & 0.2284 & 0.2353 & \textbf{0.2757}\\\hline
22 & 0.2298 & 0.2719 & 0.2480 & 0.2350 & 0.2321 & 0.2378 & 0.2291 & 0.2483 & 0.2437 & 0.2490 & \textbf{0.2808}\\\hline
27 & 0.2333 & 0.2739 & 0.2558 & 0.2423 & 0.2407 & 0.2444 & 0.2355 & 0.2523 & 0.2524 & 0.2580 & \textbf{0.2823}\\\hline
33 & 0.2390 & 0.2812 & 0.2654 & 0.2517 & 0.2469 & 0.2531 & 0.2470 & 0.2581 & 0.2619 & 0.2648 & \textbf{0.2874}\\
\hline
  \end{tabular}
  \vspace{0.3cm}
  \caption{Semantic segmentation performance in terms of mIoU when labeled data is selected using baseline active learning methods and our method on ScanNet~\cite{dai2017scannet} dataset.}
  \label{tab:res_scannet}
\end{table*}

\begin{table*}[h]
  \centering
  \small
  \begin{tabular}{|l|l|l|l|l|l|l|l|l|l|l|}
  \hline
\pbox{15cm}{\% Labeled \\ Data}&RAND&RMCDR&MCDR&ENT&CONF&CSET&MAR&MREP&\pbox{15cm}{ViewAL\\(Images)}&ViewAL\\
\hline\hline
1 & 0.0754 & 0.0797 & 0.0825 & 0.0765 & 0.0762 & 0.0778 & 0.0781 & 0.0807 & 0.0815 & 0.0802\\\hline
5 & 0.1086 & 0.1589 & 0.1250 & 0.1141 & 0.1207 & 0.1053 & 0.1159 & 0.1254 & 0.1157 & \textbf{0.1693}\\\hline
9 & 0.1310 & 0.1831 & 0.1443 & 0.1424 & 0.1387 & 0.1254 & 0.1343 & 0.1512 & 0.1496 & \textbf{0.1920}\\\hline
13 & 0.1429 & 0.1905 & 0.1659 & 0.1590 & 0.1544 & 0.1481 & 0.1478 & 0.1644 & 0.1708 & \textbf{0.2005}\\\hline
17 & 0.1564 & 0.1991 & 0.1735 & 0.1692 & 0.1616 & 0.1609 & 0.1614 & 0.1749 & 0.1750 & \textbf{0.2026}\\\hline
20 & 0.1609 & 0.1994 & 0.1802 & 0.1787 & 0.1703 & 0.1680 & 0.1673 & 0.1845 & 0.1813 & \textbf{0.2092}\\\hline
24 & 0.1660 & 0.2007 & 0.1903 & 0.1836 & 0.1796 & 0.1826 & 0.1769 & 0.1945 & 0.1925 & \textbf{0.2140}\\\hline
27 & 0.1766 & 0.2042 &  0.1947 & 0.1826 & 0.1839 & 0.1850 & 0.1777 & 0.1971 & -  & \textbf{0.2148}\\\hline
31 &0.1823 & 0.2112 & 0.2032 & 0.1960 & 0.1915 & 0.1902 & 0.1869 & 0.2019 & - & \textbf{0.2159}\\
\hline
  \end{tabular}
  \vspace{0.3cm}
  \caption{Semantic segmentation performance in terms of mIoU when labeled data is selected using baseline active learning methods and our method on Matterport3D~\cite{chang2017matterport3d} dataset.}
  \label{tab:res_matterport}
\end{table*}

\begin{table*}[h]
  \centering
  \small
  \begin{tabular}{|l|l|l|l|l|}
  \hline
\pbox{15cm}{\% Labeled \\ Data} & ViewAL(VE) & ViewAL(VE+Spx) & ViewAL(VE+Spx+MCDR) & ViewAL(VE+Spx+MCDR+VD)\\
\hline\hline
1 & 0.1004 & 0.1001 & 0.0952 & 0.0952\\\hline
6 & 0.1795 & 0.2280 & 0.2345 & \textbf{0.2365}\\\hline
12 & 0.2033 & 0.2502 & 0.2587 & \textbf{0.2663}\\\hline
17 & 0.2247 & 0.2590 & 0.2708 & \textbf{0.2757}\\\hline
22 & 0.2380 & 0.2637 & 0.2754 & \textbf{0.2807}\\\hline
27 & 0.2445 & 0.2675 & 0.2801 & \textbf{0.2822}\\\hline
33 & 0.2556 & 0.2680 & 0.2804 & \textbf{0.2873}\\
\hline
  \end{tabular}
  \vspace{0.3cm}
  \caption{Ablation Study Results. ViewAL(VE) is our method without superpixels, MC dropout, and view divergence. When superpixels are used for selection over entire images, we see significant improvements as shown by the curve ViewAL(VE+Spx). Adding MC dropout improves performance further as indicated by ViewAL(VE+Spx+MCDR). Our final method, ViewAL(VE+Spx+MCDR+VD) improves over this further by adding view divergence.}
  \label{tab:ablation}
\end{table*}

\end{document}